\documentclass[11pt]{article}
\usepackage{eacl2017}
\usepackage[backend=biber,authordate]{biblatex-chicago}
\bibliography{reference}
\usepackage{subcaption}
\usepackage{multirow}
\usepackage{graphicx}
\usepackage{times}
\usepackage{url}
\usepackage{latexsym}
\usepackage{algorithm2e}
\usepackage{tabularx}
\usepackage{booktabs}
\usepackage{supertabular}
\usepackage{float}
\usepackage[table,xcdraw]{xcolor}
\usepackage[multiple]{footmisc}
\usepackage[colorlinks = true,
            linkcolor = blue,
            citecolor =,
            hyperfootnotes=false,
            ]{hyperref}

\eaclfinalcopy

\title{Analysis of Male and Female Speakers' Word Choices in Public Speeches}

\author{Md Zobaer Hossain \\   \normalfont {\texttt{ m.z.hossain@student.rug.nl}} \\ University of Groningen \\ Groningen, The Netherlands\And
        Ahnaf Mozib Samin  \\  \normalfont {\texttt{asamin9796@gmail.com}} \\ University of Groningen \\ Groningen, The Netherlands \\
}

\begin{document}

\maketitle


\begin{abstract}
The extent to which men and women use language differently has been questioned previously. Finding clear and consistent gender differences in language is not conclusive in general, and the research is heavily influenced by the context and method employed to identify the difference. In addition, the majority of the research was conducted in written form, and the sample was collected in writing. Therefore, we compared the word choices of male and female presenters in public addresses such as TED lectures. The frequency of numerous types of words, such as parts of speech (POS), linguistic, psychological, and cognitive terms were analyzed statistically to determine how male and female speakers use words differently. Based on our data, we determined that male speakers use specific types of linguistic, psychological, cognitive, and social words in considerably greater frequency than female speakers.
\end{abstract}

\section{Introduction}
Previous social science research suggests that men are more prone to use language for individualism, whilst women are more likely to use language for social verbal contact \autocite{brownlow2003gender,colley2004style,mulac2001empirical}. However, according to \textcite{thomson2001predicting}, there are no substantial differences between men and women when it comes to e-mail communication. More recently, \textcite{newman2008gender} examined gender differences in language use by analyzing over 14,000 texts from 70 distinct research. They discovered that women made more use of psychological and social terminology, whilst men made more use of impersonal characteristics and subjects. Finding clear and consistent gender differences in language, in general, is not definitive, and the research is largely influenced by the context and technique used to find the difference. Furthermore, the majority of the research has been conducted in texts, with the sample being collected in written form. As a result, we investigated a comparison of male and female presenters' word choices in public speeches such as TED talks. For this investigation, we formulate the following research questions:

\begin{itemize}
    \item R1: Are there substantial disparities in word selections between male and female speakers?
    \item R2: If they do differ, is there any circumstance in which gender disparities in language are significantly more pronounced?
\end{itemize}

We utilized a log odds ratio to assess how word usage varies by gender. Additionally, a MANOVA analysis was performed using the frequency of several categories of words, such as parts of speech (POS), linguistic, psychological, or cognitive terms to find out how male and female speakers utilize words differently. Based on statistical analysis, we discovered that male speakers employ particular types of linguistic, psychological, cognitive, or social words in significantly higher numbers than female speakers. 

Furthermore, we trained a computational model based on a state-of-the-art Transformer architecture to investigate whether such a model can identify the gender given the transcriptions of public speeches \autocite{liu2019roberta}. We hypothesize that if our model can identify the gender with high accuracy, then there must be certain distinguishable elements between male and female speech. The empirical evaluation suggests such evidence of differences between male and female speech.

\section{Related Works}
Researchers have been trying to find the difference in language use by males and females. For instance, \textcite{mehl2003sounds} found that females were more likely to employ the first-person singular pronouns. Furthermore, females have been observed to employ more intensive adverbs, conjunctions such as \textit{but}, and modal auxiliary verbs such as \textit{could} \autocite{biber1998corpus, mehl2003sounds, mulac2001empirical}. Males, on the other hand, have been observed to swear more frequently, use longer words, include more articles, and make more geographical references \autocite{gleser1959relationship, mehl2003sounds, erickson1978speech}. Despite the existence of distinct stereotypes, the findings corresponding to the usage of emotional words are not consistent. For example, many studies have revealed that females use emotional words more frequently than males \autocite{mulac1986linguistic,thomson2001predicting}. However, \textcite{mulac2000female} discovered precisely the opposite in their research of managers offering criticism in role-plays.

More recently, \textcite{newman2008gender} examined gender differences in language use by analyzing over 14,000 texts from 70 distinct research. They discovered that women made more use of psychological and social terminology, whilst men made more use of impersonal characteristics and subjects. However, evaluating spoken data is a relatively infrequent occurrence in this area. \textcite{newman2008gender} investigated transcripts from talk shows, although their spoken sample included the transcripts of 168 male and 219 female speakers. As a result, we assessed and analyzed a larger dataset, which comprised of transcripts from TED talks.

\section{Dataset}
The TEDx\footnote{\url{https://github.com/go2chayan/TEDTalk\_Analytics}}  open dataset is used in this paper for analysis. It consists of 2,007 talks that were published prior to September 2016 and does not include any videos that involve music, dancing, or entertainment performances. It also includes metadata such as title, short description, speaker's name, video duration, total views, viewer ratings in 14 different categories, keywords, and publication date, in addition to the transcripts. The transcripts are originally collected in the English language. Each presentation is approximately 14 minutes in length, and each transcript has approximately $\approx 2.6K$ words on average. However, there is no information about the speakers' gender, origin, or profession in this dataset. As a result, we will augment the dataset for each speaker's information by leveraging information from Wikipedia\footnote{\url{https://www.wikidata.org/wiki/Wikidata:TED/TED\_speakers}} to include gender and origin. In our study, we will use the transcripts of speeches to analyze the gender difference in language.

Several speakers' gender and country of origin were unknown to us. Moreover, in our experiment, we only retained one speech from each speaker, resulting in a total of 941 speeches from the TEDx open dataset, with 643 male speakers and 298 female speakers. 

\section{Method}

\subsection{Weighted Log Odds ratio}

To measure how the usage of words differs across different texts log odds ratio can be used. Odds ratio (OR) is a statistic that measures how strong the connections between two events, A and B, is. The odds ratio can be described as the ratio of the probability of occurrence of A in the presence of B to the probability of occurrence of A in the absence of B. The logarithm of the odds ratio makes the measure symmetrical in terms of event order. Nevertheless, the log odds ratio by itself can not quantify sampling heterogeneity because not all features or words are counted equally. Hence, it is difficult to assess whether the log odds ratio i.e the difference of the logits of the probabilities is actually meaningful. \textcite{monroe2008fightin} have proposed a method of Weighted Log Odds ratio in which the log odds ratio is weighted by a prior estimate from the data according to an empirical bayes approach. In our study, we have utilized this technique to analyze the difference in usage of unigrams and bigrams by male and female speakers.

\subsection{Linguistic Inquiry and Word Count (LIWC)}

Throughout history, scholars have acquired increasingly definitive evidence that the words we choose have immense psychological significance \autocite{roberts2020text, weintraub1989verbal, green1967reviews, bucci1981language}. Individual differences in the use of function words reflect individual variances in how they think about and react to the environment \autocite{newman2008gender,pennebaker2003psychological}.   A text analysis program called Linguistic Inquiry and Word Count (LIWC) \autocite{pennebaker2001linguistic} was developed in order to provide an efficient method for analyzing different sentimental, psychological, and functional components present in individuals' verbal and written speech samples. In our experiment, we used the LIWC lexicon to count the number of words from various categories that were present in the transcripts of the male and female speakers, respectively. The LIWC dictionary, which was employed in this study, was made up of 72 categories and 6549 words, with several terms belonging to multiple categories at the same time. We used multivariate analysis of variance (MANOVA) and univariate analysis of variance (ANOVA) to examine the effect of gender on the use of LIWC words as a whole group and as an individual category, respectively. In order to eliminate the problem of multiple comparisons in the ANOVA, we changed the alpha level according to Bonferroni's correction method \footnote{\url{https://en.wikipedia.org/wiki/Multiple_comparisons_problem}}\footnote{\url{https://en.wikipedia.org/wiki/Bonferroni_correction}}.

\subsection{Part of Speech (POS)}

Though LIWC categories contain several parts of speech (POS) tags, we conducted a separate analysis of POS tags for the purpose of improving our assessment. We did not take the context of the sentence into consideration when counting the POS tags found in LIWC dictionaries. As a result, we used a more comprehensive POS tagger, spacy\footnote{\url{https://spacy.io/usage/linguistic-features}}, to determine the frequency of POS tags in a given transcript. Each token was assigned a unique POS tag from a list of 19 available tags\footnote{\url{https://machinelearningknowledge.ai/tutorial-on-spacy-part-of-speech-pos-tagging/}}. We used the frequency of POS tags to conduct a MANOVA analysis to determine the influence of gender on all POS tags combined. Additionally, we used the ANOVA test to determine individual differences.

\subsection{Computational Modeling}
Previous studies demonstrated the effectiveness of machine learning (ML) techniques in identifying gender from texts \autocite{cheng2011author, sboev2018automatic}. The hypotheses underlying these experiments are that if a trained ML classifier can identify the gender with a reasonable degree of accuracy, then there should be some distinguishable features in male and female writing for which the model could perform well. However, no previous research leveraged the ML algorithms to investigate the public speeches of male and female speakers, to the best of our knowledge. In this work, we analyzed the probable gender disparity in the public speech domain by utilizing RoBERTa, a state-of-the-art large-scale language model pre-trained with the transformer architecture in a self-supervised way on a large amount of data \autocite{liu2019roberta}. We finetuned this model using our TEDx dataset. 

We performed 5-fold cross-validation to ensure the reliability of our evaluation. Thus, we split the TEDx dataset into five equal sets, merging four of the sets into the training set and keeping the remaining one as the evaluation set, iteratively. Subsequently, the models were trained five times and the evaluation set accuracy was averaged. Since the number of speeches of the male speakers is considerably higher than that of female speakers in our dataset, we upsampled the female speakers' texts in the training set to address the imbalance in the dataset. After upsampling, our training set and evaluation set contained approximately 82\% and 18\% data, respectively, in each iteration. Finally, we measured accuracy and macro F1-score of our model to determine if there was any indication of gender disparity.

\section{Statistical Analysis}
\subsection{Word frequency analysis}

\begin{table}[h]
\centering
\begin{tabular}{lllll}
\toprule
Gender & mean    & min & max  & SD      \\
\midrule
Male   & 2778.62 & 493 & 6914 & 1113.68 \\
Female & 2419.03 & 490 & 4803 & 1004.97 \\
\bottomrule
\end{tabular}
\caption{Mean, minimum, maximum, and standard deviation (SD) of word counts in male and female transcripts.}
\label{tab:word}
\end{table}

 \begin{figure}[h]
    \centering
    \includegraphics[width=\linewidth]{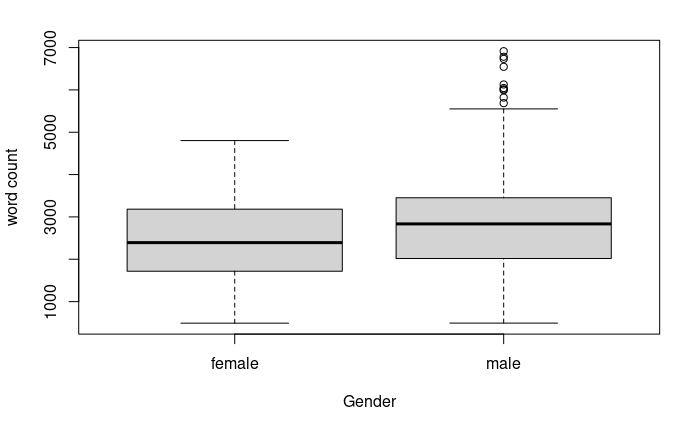}
    \caption{Comparison of the distribution of word counts in male and female speakers' transcripts with box plots.}
    \label{fig:word-bp}
\end{figure}

 \begin{figure}[h]
    \centering
    \includegraphics[width=\linewidth]{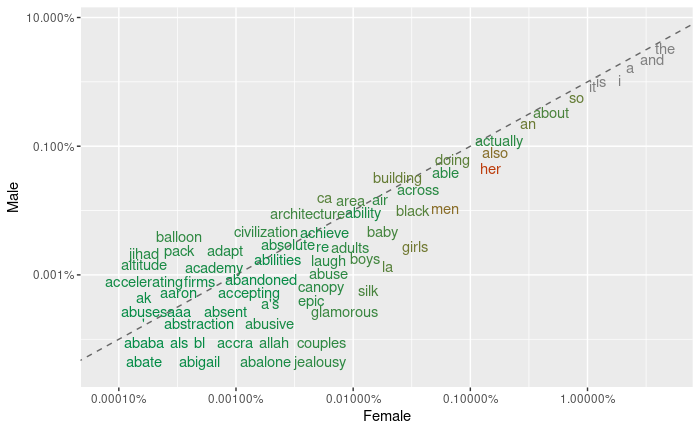}
    \caption{A comparison of the word frequencies in male and female speakers' transcripts.}
    \label{fig:word-fa}
\end{figure}

 We used statistical analysis and significance tests to look for potential disparities in word selections. To begin, we examined the average word usage in male and female speakers' TED speeches. Male speakers, on average, used more words than female ones (Table \ref{tab:word}). However, the boxplot (Figure \ref{fig:word-bp}) indicated that descriptive statistics on word usage were relatively comparable across genders, with the exception that the distribution of male speakers' word counts contained several outliers. We produced Figure \ref{fig:word-fa} to compare the male and female speakers' word frequencies. The words that ran parallel to the lines in these plots occurred frequently in both sets of texts, such as \textit{academy}, \textit{absolute}, and \textit{doing}. On the other hand, words that occurred frequently beyond the line appeared more commonly in one collection of texts than in another, for example, \textit{balloon} for males and \textit{jealousy} for females. Additionally, we conducted a correlation analysis to determine the relationship between these word frequencies. The Pearson $r$ correlation analysis demonstrated a positive association between male and female speakers' word frequencies ($r(17472) = 0.995$, $p < .001$, two-tailed, 95 percent confidence interval $[0.995, 0.996]$). Given that $r^2 \approx 99.0$, the effect size is regarded to be large. As a result, we may conclude that the association is significant and that the word usage is similar in both transcripts.

\begin{table*}[h]
\centering
\begin{tabular}{llcllc}
\toprule
   Category    & \multicolumn{1}{c}{F Value} & \multicolumn{1}{c}{Num  DF}& \multicolumn{1}{c}{Den  DF} & \multicolumn{1}{c}{Pr \textgreater F} &  \multicolumn{1}{c}{Effect Size $(\eta^2)$}  \\
      \midrule
\textbf{With Outliers}       & & & & &\\
\midrule
LIWC    & 4.062   & 43     & 897    & 6.9e-16 & 0.16        \\
POS     & 8.456  & 14     & 926    & 2.2e-16  & 0.11         \\

\midrule
\textbf{Without Outliers}       & & & & \\
\midrule
LIWC    & 3.622  &   43  &  798 & 4.1e-13   & 0.16    \\
POS     & 8.152  &   14  &  887 & 2.2e-16     & 0.11       \\

\bottomrule

\end{tabular}

\caption{Main Effects of gender on the overall group in LIWC and POS tags with and without outlier observations.}
\label{tab:manova}
\end{table*}

\subsection{Weighted Log Odds ratio}

\begin{figure}[h]
    \centering
    \includegraphics[width=\linewidth]{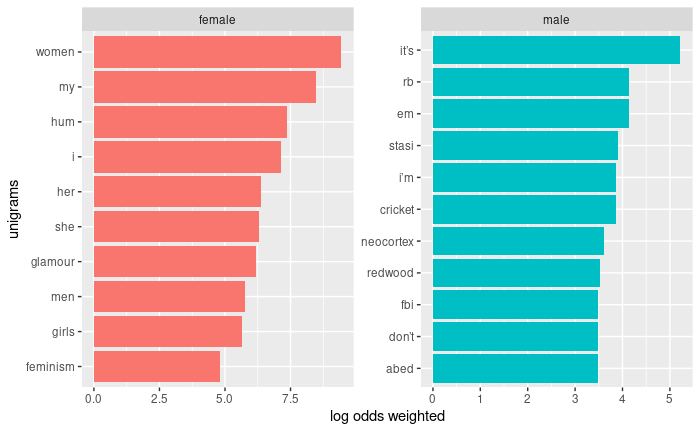}
    \caption{The ten unigrams from male and female transcripts with the most weighted log-odds.}
    \label{fig:unigram}
\end{figure}

\begin{figure}[h]
    \centering
    \includegraphics[width=\linewidth]{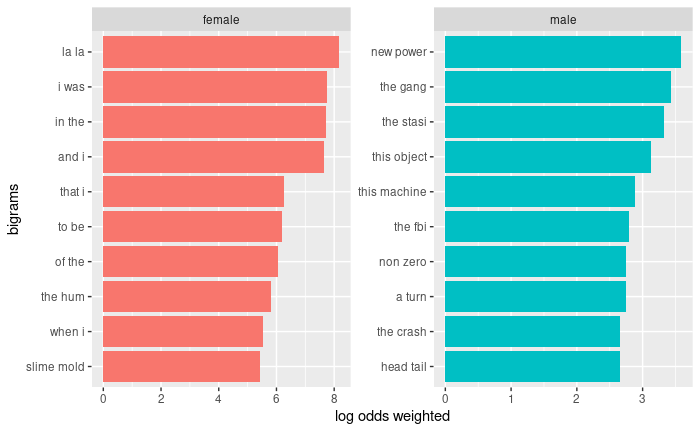}
    \caption{The ten bigrams from male and female transcripts with the most weighted log-odds.
}
    \label{fig:bigram}
\end{figure}

The difference in the use of unigrams and bigrams by male and female speakers was analyzed using the weighted log odds ratio technique. While prior analysis revealed a similarity in word frequencies between male and female transcripts, none of the unigrams from the top ten most weighted log odds ratios match between male and female transcripts (Figure \ref{fig:unigram}). The same effect was observed with bigrams' weighted log odds ratio (Figure \ref{fig:bigram}). This indicated that there were some distinctions in the way the word was used by male and female speakers.

\subsection{MANOVA}

\begin{figure*}[h]
\begin{subfigure}{.5\textwidth}
\centering
\includegraphics[width=\columnwidth]{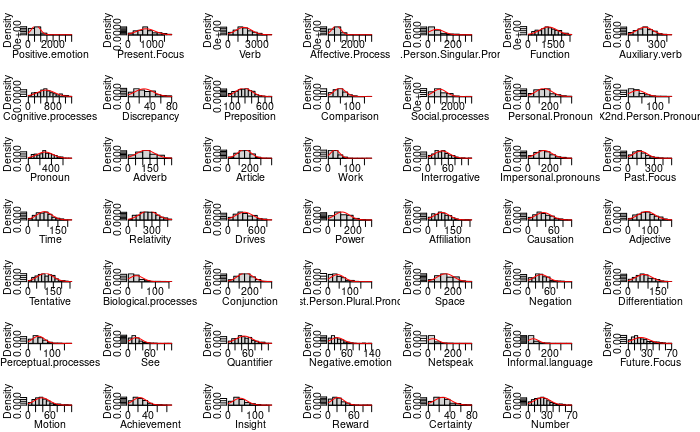}
\caption{Female}
\end{subfigure}
\begin{subfigure}{.5\textwidth}
\centering
    \centering
    \includegraphics[width=\columnwidth]{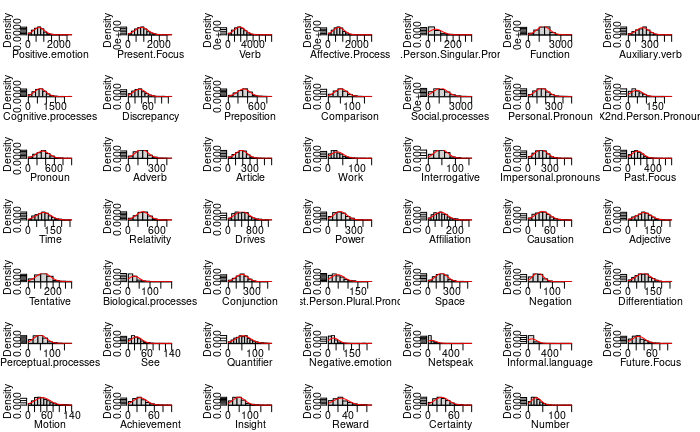}
    \caption{Male}
\end{subfigure}
    \caption{Probabilities of word counts with a normal distribution curve plotted over the data in every LIWC categories.}
    \label{fig:liwc-curve}
\end{figure*}

\begin{figure*}[h]
\begin{subfigure}{.5\textwidth}
\centering
\includegraphics[width=\columnwidth]{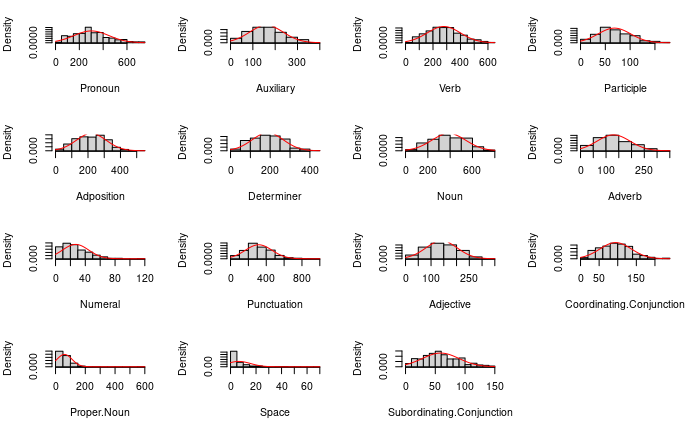}
\caption{Female}
\end{subfigure}
\begin{subfigure}{.5\textwidth}
\centering
    \centering
    \includegraphics[width=\columnwidth]{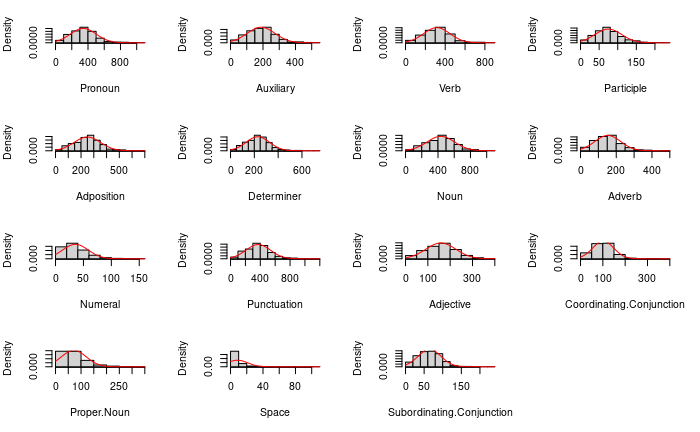}
    \caption{Male}
\end{subfigure}
    \caption{Probabilities of word counts with a normal distribution curve plotted over the data in every POS tags.}
    \label{fig:pos-curve}
\end{figure*}

Prior to the experiment, we prepared two data frames by counting the number of words from each LIWC category and POS tag that were included in the transcripts of male and female speakers. To avoid ambiguity, we shall refer to them as the LIWC and POS datasets in the rest of the paper. Our new datasets consisted of multiple dependent variables. For example, LIWC contains $72$ categories whereas POS tags consist of $19$ categories. However, we did not include all LIWC and POS tags. We discarded variables with extremely low mean values. For instance, we eliminated categories from the LIWC dataset with a mean of less than $20$. Thus, after removing the $24$ variables, we found $48$ categories with a relatively higher mean. We deleted tags with a mean of less than $10$ from the POS dataset. Finally, we obtained the POS dataset which contained $16$ tags. Since both datasets had multiple dependent variables, we opted to perform the MANOVA test instead of ANOVA or other statistical tests such as the Student t-test. The Multivariate Analysis Of Variance (MANOVA) is an ANOVA with two or more continuous outcome (or response) variables. The one-way MANOVA tests simultaneously statistical differences for multiple response variables by one grouping variable. The basic idea behind MANOVA is to create a new variable from the linear combination of all dependent variables and compare this new variable of different groups with mean values. However, MANOVA assumes the presence of certain characteristics about the data. These characteristics are:

\begin{itemize}
    \item \textbf{Independence of the observations:} Each observation should be assigned to a single group. There should not be any correlation between the observations in the different groups. This criterion was rigorously evaluated for the LIWC and POS datasets.
    
    \item \textbf{Outliers:} There should not be any outliers in either the univariate or multivariate data. We employed the Mahalanobis\footnote{\url{https://en.wikipedia.org/wiki/Mahalanobis_distance}} distance to identify multivariate outliers. The distance parameter indicated how far an observation was from the cluster's center, taking into account the cluster's shape (covariance). However, in other circumstances, deleting outliers could have a detrimental effect; therefore, we chose to report findings both with and without outliers. We discovered $99$ outlier observations in the LIWC dataset, compared to $39$ in the POS dataset. Additionally, we used the \textit{rstatix}\footnote{\url{https://cran.r-project.org/web/packages/rstatix/rstatix.pdf}} package to calculate the Mahalanobis distance in R.
    
    \item \textbf{Normality:} Because we had more than $200$ samples for each group, we determined the dataset's normality using the histogram and normality curves for each dependent variable across distinct groups (Male, Female). The variables whose histograms differed significantly from a typical normal distribution were excluded. Additionally, MANOVA is fairly resilient to deviations from normality; hence, this assumption should have little effect on the final outcome. 
    
    We discovered several categories in the LIWC dataset that had a probability density that deviated from the normal distribution, including Biological processes, Negative Emotion, Informal language, and Netspeak (Figure \ref{fig:liwc-curve}). On the other hand, the POS dataset's Space tag is the only one that deviated from the normal distribution (Figure \ref{fig:pos-curve}). We eliminated these variables before doing the MANOVA test.
    
    \item \textbf{Absence of multicollinearity:} Correlations between dependent variables should preferably be moderate, not excessive. Correlation coefficients greater than $0.9$ indicate multicollinearity, which is troublesome for MANOVA. When the association was confirmed to be extremely high, the variables were discarded. We discovered a strong correlation between the \textit{1st person singular} and \textit{1st person plural pronoun} categories. As a result, the \textit{1st person singular pronoun} category was removed from the LIWC dataset. However, no POS tag was shown to be substantially associated with any other POS tags. 
    
    \item \textbf{Linearity:} For each group, the pairwise connection between the dependent variables should be linear to perform MANOVA. We created scatter plots for each pair of dependent variables for two distinct groups and determined whether or not the relationship is linear. Our study revealed the presence of a linear relationship among variables \footnote{There were just too many plots for the paper to address. Hence , we excluded them.}.

    \item \textbf{Homogeneity of variances:} The one-way MANOVA implies that the variances between groups should be identical for each dependent variable. This can be verified using Levene's test for the variance of equality. Additionally, we utilized the \textit{rstatix} package to do Levene's test in this case. Ten categories in the LIWC dataset failed Levene's test. On the other hand, this Levene's test fails in only two POS tags (\textit{Determinant, Number}). However, by lowering the alpha level (statistical significance), we could still do the MANOVA test, and additional post-hoc tests could complement the MANOVA result. 

    \item \textbf{Homogeneity of variance-covariance matrices:} To verify this assumption, we employed the Box's M test\footnote{\url{https://en.wikipedia.org/wiki/Box\%27s_M_test}} as implemented in the \textit{rstatix} package. In both datasets, the data violated the assumption that variance-covariance matrices are homogeneous. However, in the event of a balanced dataset with an equal number of entries for each group, violating the homogeneity of variance-covariance matrices does not cause significant problems. However, because the quantity of transcripts for male speakers was greater, we employed Pillai's multivariate statistic rather than Wilks' in the MANOVA test to minimize the violation of this criteria.

\end{itemize}

A one-way multivariate analysis of variance was performed to determine the effect of gender on LIWC categories. There was a statistically significant difference between the gender on the combined dependent variables, $F(43, 897) = 4.062, p < 0.0001$. The effect size was found large ($\eta^2 = 0.16$). Since the homogeneity of variance assumption was violated, we performed Welch ANOVA for the post-hoc analysis. Follow-up univariate Welch ANOVAs, using a Bonferroni adjusted alpha level of $0.002$, showed that there was a statistically significant difference in LIWC categories between male and female speakers except for \textit{Conjunction}, \textit{Past Focused}, \textit{Insight}, \textit{Negation}, \textit{Pronoun}, \textit{1st person plural pronoun}, and \textit{Work} (Table \ref{tab:liwc-example}).

Furthermore, another one-way multivariate analysis of variance was performed to determine the effect of gender on POS tags. There was a statistically significant difference between the gender on the combined dependent variables, $F(14, 926) = 8.4557, p < 0.0001$. The effect size was medium ($\eta^2 = 0.11$). Follow-up univariate Welch ANOVAs, using a Bonferroni adjusted alpha level of $0.004$, showed that there was a statistically significant difference in POS tags between male and female speakers except for \textit{coordinating conjunction, particle, subordinating conjunction} (Table \ref{tab:pos-welch}). 

Due to the presence of outliers in the dataset, we performed MANOVA on the remaining observations after removing the outliers. The significance levels and effect sizes, on the other hand, stayed the same (Table \ref{tab:manova}).

\section{Discussion}
According to preliminary descriptive statistics and frequency analyses, the word choices appeared to be rather similar. However, further study revealed significant discrepancies in word choices between male and female speakers. Both with and without outliers, the main effects of gender on the overall group were significant in both LIWC and POS tags, with $p$ values near to zero. As a result, it was not necessary to reduce the alpha value due to the violation of the assumption of variance homogeneity. Additionally, it was reasonable to presume that the effect was significant. The effect size for MANOVA with the LIWC dataset was found to be large, implying that gender had a significant effect on the LIWC categories overall. Gender, on the other hand, had a medium effect on POS tags.

Post hoc analysis revealed that the speaker's gender had no effect on several LIWC and POS categories. While prior research indicated a difference in females' use of first-person pronouns and conjunctions, our analysis revealed no significant effect of gender on these categories of terms. Additionally, the mean values for all categories were higher for male speakers, implying that the categories with significantly different mean values were more frequently utilized by male speakers. This also contradicted a prior study that indicated that certain categories were more prevalent in female writings than in male writings \autocite{pennebaker2001linguistic,biber1998corpus, mulac2001empirical}.

With regard to our computational modeling approach, RoBERTa could identify the gender with 73.22\% accuracy and 64\% macro F1-score given the speech transcriptions as inputs. This result suggests that there are subtle distinguishable characteristics in the male and female speeches, although these characteristics are not particularly prominent. However, the performance of such computational models largely depends on the size of the training data, so it is likely that the accuracy and F1-score would increase with more data.


\section{Conclusion}
In this study, we looked into two main research questions. 
\begin{itemize}
    \item R1: Are there substantial disparities in word selections between male and female speakers?
    \item R2: If they do differ, is there any circumstance in which gender disparities in language are significantly more pronounced?
\end{itemize}

Based on our research, we found that male speakers used more linguistic, psychological, cognitive, or social words than female speakers. However, our conclusions may not be applicable in the real world unless we thoroughly investigate all important variables, such as the TED Talk's topic, the speakers' educational backgrounds, and whether the speaker was an L1 or L2 English learner. 
Furthermore, to determine if models trained on the TedX dataset accurately predict the speaker's gender, advanced Transformer models such as RoBERTa has been used. Our hypothesis is that if the model can accurately identify gender, there is a considerable difference in how male and female speakers utilize words. This experiment implies that there are differences between male and female speech, however, these differences are not extremely pronounced.

\printbibliography

@article{brownlow2003gender,
  title={Gender-linked linguistic behavior in television interviews},
  author={Brownlow, Sheila and Rosamond, Julie A and Parker, Jennifer A},
  journal={Sex Roles},
  volume={49},
  number={3},
  pages={121--132},
  year={2003},
  publisher={Springer}
}

@article{colley2004style,
  title={Style and content in e-mails and letters to male and female friends},
  author={Colley, Ann and Todd, Zazie and Bland, Matthew and Holmes, Michael and Khanom, Nuzibun and Pike, Hannah},
  journal={Journal of Language and Social Psychology},
  volume={23},
  number={3},
  pages={369--378},
  year={2004},
  publisher={Sage Publications Sage CA: Thousand Oaks, CA}
}

@article{thomson2001predicting,
  title={Predicting gender from electronic discourse},
  author={Thomson, Rob and Murachver, Tamar},
  journal={British Journal of Social Psychology},
  volume={40},
  number={2},
  pages={193--208},
  year={2001},
  publisher={Wiley Online Library}
}

@article{newman2008gender,
  title={Gender differences in language use: An analysis of 14,000 text samples},
  author={Newman, Matthew L and Groom, Carla J and Handelman, Lori D and Pennebaker, James W},
  journal={Discourse processes},
  volume={45},
  number={3},
  pages={211--236},
  year={2008},
  publisher={Taylor \& Francis}
}

@article{pennebaker2003psychological,
  title={Psychological aspects of natural language use: Our words, our selves},
  author={Pennebaker, James W and Mehl, Matthias R and Niederhoffer, Kate G},
  journal={Annual review of psychology},
  volume={54},
  number={1},
  pages={547--577},
  year={2003},
  publisher={Annual Reviews 4139 El Camino Way, PO Box 10139, Palo Alto, CA 94303-0139, USA}
}

@book{roberts2020text,
  title={Text analysis for the social sciences: methods for drawing statistical inferences from texts and transcripts},
  author={Roberts, Carl W},
  year={2020},
  publisher={Routledge}
}

@book{weintraub1989verbal,
  title={Verbal behavior in everyday life.},
  author={Weintraub, Walter},
  year={1989},
  publisher={Springer Publishing Co}
}

@article{pennebaker2001linguistic,
  title={Linguistic inquiry and word count: LIWC 2001},
  author={Pennebaker, James W and Francis, Martha E and Booth, Roger J},
  journal={Mahway: Lawrence Erlbaum Associates},
  volume={71},
  number={2001},
  pages={2001},
  year={2001}
}

@article{monroe2008fightin,
  title={Fightin'words: Lexical feature selection and evaluation for identifying the content of political conflict},
  author={Monroe, Burt L and Colaresi, Michael P and Quinn, Kevin M},
  journal={Political Analysis},
  volume={16},
  number={4},
  pages={372--403},
  year={2008},
  publisher={Cambridge University Press}
}

@article{mehl2003sounds,
  title={The sounds of social life: a psychometric analysis of students' daily social environments and natural conversations.},
  author={Mehl, Matthias R and Pennebaker, James W},
  journal={Journal of personality and social psychology},
  volume={84},
  number={4},
  pages={857},
  year={2003},
  publisher={American Psychological Association}
}

@book{biber1998corpus,
  title={Corpus linguistics: Investigating language structure and use},
  author={Biber, Douglas and Conrad, Susan and Reppen, Randi},
  year={1998},
  publisher={Cambridge University Press}
}

@article{mulac2001empirical,
  title={Empirical support for the gender-as-culture hypothesis: An intercultural analysis of male/female language differences},
  author={Mulac, Anthony and Bradac, James J and Gibbons, Pamela},
  journal={Human Communication Research},
  volume={27},
  number={1},
  pages={121--152},
  year={2001},
  publisher={Oxford University Press}
}

@article{gleser1959relationship,
  title={The relationship of sex and intelligence to choice of words: a normative study of verbal behavior.},
  author={Gleser, Goldine C and Gottschalk, Louis A and John, Watkins},
  journal={Journal of Clinical Psychology},
  year={1959},
  publisher={John Wiley \& Sons}
}

@article{mulac1986linguistic,
  title={Linguistic contributors to the gender-linked language effect},
  author={Mulac, Anthony and Lundell, Torborg Louisa},
  journal={Journal of Language and Social Psychology},
  volume={5},
  number={2},
  pages={81--101},
  year={1986},
  publisher={MULTILINGUAL MATTERS LTD. Bank House, 8a Hill Rd, Clevedon, Avon, England~…}
}

@article{mulac2000female,
  title={Female and male managers’ and professionals’ criticism giving: Differences in language use and effects},
  author={Mulac, Anthony and Seibold, David R and Farris, Jennifer Lee},
  journal={Journal of Language and Social Psychology},
  volume={19},
  number={4},
  pages={389--415},
  year={2000},
  publisher={Sage Publications Sage CA: Thousand Oaks, CA}
}

@article{erickson1978speech,
  title={Speech style and impression formation in a court setting: The effects of “powerful” and “powerless” speech},
  author={Erickson, Bonnie and Lind, E Allan and Johnson, Bruce C and O'Barr, William M},
  journal={Journal of experimental social psychology},
  volume={14},
  number={3},
  pages={266--279},
  year={1978},
  publisher={Elsevier}
}

@article{bucci1981language,
  title={The language of depression},
  author={Bucci, Wilma and Freedman, Norbert},
  journal={Bulletin of the Menninger Clinic},
  volume={45},
  number={4},
  pages={334},
  year={1981},
  publisher={Menninger Foundation.}
}

@article{green1967reviews,
  title={Reviews: Stone, Philip J., Dunphy, Dexter C., Smith, Marshall S., and Ogilvie Daniel M. The General Inquirer: A Computer Approach to Content Analysis. Cambridge, Mass.: MIT Press, 1966. Pp. xx+ 651. 7.95.},
  author={Green Jr, Bert F},
  journal={American Educational Research Journal},
  volume={4},
  number={4},
  pages={397--398},
  year={1967},
  publisher={Sage Publications}
}

@article{cheng2011author,
  title={Author gender identification from text},
  author={Cheng, Na and Chandramouli, Rajarathnam and Subbalakshmi, KP},
  journal={Digital investigation},
  volume={8},
  number={1},
  pages={78--88},
  year={2011},
  publisher={Elsevier}
}

@article{sboev2018automatic,
  title={Automatic gender identification of author of Russian text by machine learning and neural net algorithms in case of gender deception},
  author={Sboev, Alexander and Moloshnikov, Ivan and Gudovskikh, Dmitry and Selivanov, Anton and Rybka, Roman and Litvinova, Tatiana},
  journal={Procedia computer science},
  volume={123},
  pages={417--423},
  year={2018},
  publisher={Elsevier}
}

@article{liu2019roberta,
  title={Roberta: A robustly optimized bert pretraining approach},
  author={Liu, Yinhan and Ott, Myle and Goyal, Naman and Du, Jingfei and Joshi, Mandar and Chen, Danqi and Levy, Omer and Lewis, Mike and Zettlemoyer, Luke and Stoyanov, Veselin},
  journal={arXiv preprint arXiv:1907.11692},
  year={2019}
}

\section*{Appendices}
The code for this study can be found at \url{https://github.com/Rowan1697/Corpus}. 

\onecolumn

\bottomcaption{The mean and standard deviation (SD) of the word count across 14 POS tags in male and female transcripts. Additionally, each tag includes example words. In the highlighted tags, there was no statistically significant difference in mean between male and female speakers.}

\label{tab:pos-welch}
\tablehead
{\toprule
                          &                           & \multicolumn{2}{c}{Mean} & \multicolumn{2}{c}{SD} \\
\multirow{-2}{*}{POS tag} & \multirow{-2}{*}{Example} & male        & female     & male       & female    \\
\midrule}
\begin{center}
\begin{supertabular*}{\textwidth}{llllll}

Adjective                 & good;extraordinary               & 158.299 & 141.181 & 67.238 & 61.377 \\
Adposition                & to;in                    & 249.386 & 220.208 & 103.412 & 92.442 \\
Adverb                    & slowly;carefully               & 152.644 & 129.426 & 72.213 & 64.537 \\
Auxiliary                 & is;has                   & 188.171 & 160.809 & 85.309 & 75.905 \\
\rowcolor[HTML]{9AFF99} 
Coordinating Conjunction  & and;but                 & 101.560 & 94.342 & 47.161 & 42.283 \\
Determiner                & a;an;the                  & 229.258 & 187.430 & 97.982 & 79.489 \\
Noun                      & cat;boy                & 435.264 & 387.389 & 170.611 & 153.239 \\
Numeral                   & one;two                  & 35.661 & 26.970 & 22.893 & 17.625 \\
\rowcolor[HTML]{9AFF99} 
Participle                & 's,not                   & 74.655 & 68.372 & 35.751 & 32.753 \\
Pronoun                   & he;she                   & 342.253 & 301.594 & 158.787 & 147.267 \\
Proper Noun               & John;Marshmello             & 71.723 & 60.225 & 48.804 & 50.571 \\
Punctuation               & :,                       & 372.264 & 315.503 & 163.629 & 150.134 \\
\rowcolor[HTML]{9AFF99} 
Subordinating Conjunction & if;that                   & 64.675 & 59.064 & 30.375 & 30.636 \\
Verb                      & sleep,kick               & 315.361 & 278.899 & 132.203 & 120.792 \\
\bottomrule
\end{supertabular*}
\end{center}

\bottomcaption{The mean and standard deviation (SD) of the word count in male and female transcripts across 43 LIWC categories. Additionally, examples of words from each category are provided. There was no statistically significant difference in mean between male and female speakers in the highlighted categories.}
\label{tab:liwc-example}
\tablehead
{\toprule
\multicolumn{1}{c}{}                           & \multicolumn{1}{c}{}                          & \multicolumn{2}{c}{Mean}                              & \multicolumn{2}{c}{SD}                                \\
\multicolumn{1}{c}{\multirow{-2}{*}{Category}} & \multicolumn{1}{c}{\multirow{-2}{*}{Example}} & \multicolumn{1}{c}{male} & \multicolumn{1}{c}{female} & \multicolumn{1}{c}{male} & \multicolumn{1}{c}{female} \\
\midrule}
\begin{center}
\begin{supertabular*}{\linewidth}{llllll}
\multicolumn{6}{l}{\textbf{Linguistic Dimension}}                                                                                                                                                              \\
\midrule
Function                                       & whence;we                                     & 1,391.372                & 1,206.161                  & 577.947                  & 520.842                    \\
Pronoun                                        & it;you                                        & 333.667                  & 289.195                    & 149.261                  & 137.729                    \\
\rowcolor[HTML]{9AFF99} 
Personal Pronoun                               & them;thoust                                   & 155.278                  & 144.114                    & 74.642                   & 74.268                     \\
\rowcolor[HTML]{9AFF99} 
1st Person Plural Pronoun                      & ourselves;let's                               & 48.229                   & 42.460                     & 33.427                   & 30.927                     \\
2nd Person Pronoun                             & you'll;your                                     & 46.482                   & 33.899                     & 32.573                   & 27.819                     \\
Impersonal pronouns                            & who's;anything                                & 178.171                  & 144.903                    & 85.001                   & 74.170                     \\
Article                                        & the;an;a                                      & 183.625                  & 148.503                    & 80.111                   & 62.264                     \\
Preposition                                    & after;respecting                              & 319.034                  & 282.356                    & 131.641                  & 118.568                    \\
Auxiliary verb                                 & thats;it'd                                    & 199.135                  & 171.299                    & 86.718                   & 81.003                     \\
Adverb                                         & ever;especially                               & 140.802                  & 120.990                    & 65.720                   & 61.395                     \\
\rowcolor[HTML]{9AFF99} 
Conjunction                                    & when;how're                                    & 171.439                  & 155.970                    & 75.201                   & 70.053                     \\
\rowcolor[HTML]{9AFF99} 
Negation                                       & don't;shan't                                  & 40.003                   & 35.450                     & 21.833                   & 20.218                     \\
\midrule
\multicolumn{6}{l}{\textbf{Other Grammar}}                                                                                                                                                                     \\
\midrule
Verb                                           & managed;surfs                                 & 1,856.532                & 1,612.326                  & 908.031                  & 818.422                    \\
Adjective                                      & equal;scariest                                & 94.953                   & 83.849                     & 41.491                   & 39.073                     \\
Comparison                                     & fattiest;healthier                            & 53.070                   & 47.195                     & 25.455                   & 24.738                     \\
Interrogative                                  & how's;how                                    & 47.051                   & 41.178                     & 24.405                   & 22.582                     \\
Number                                         & four;tenth                                    & 27.804                   & 22.607                     & 16.566                   & 12.710                     \\
Quantifier                                     & sample;each                                    & 55.467                   & 47.396                     & 27.354                   & 24.738                     \\
\midrule
\multicolumn{6}{l}{\textbf{Psychological Processes}}                                                                                                                                                           \\
\midrule
Affective Process                              & friendliest;gentler                           & 731.543                  & 643.785                    & 372.199                  & 354.008                    \\
Positive emotion                               & fearless;proud                                   & 695.762                  & 614.268                    & 352.810                  & 342.799                    \\
Social processes                               & talking;excuse                                 & 1,136.462                & 924.963                    & 651.324                  & 538.136                    \\
Cognitive processes                            & couldn't;needn't                               & 660.476                  & 568.500                    & 316.392                  & 282.146                    \\
\rowcolor[HTML]{9AFF99} 
Insight                                        & wiser;knowledge                                & 48.602                   & 43.896                     & 28.269                   & 25.597                     \\
Causation                                      & comply;provoke                                 & 39.515                   & 34.919                     & 19.743                   & 18.150                     \\
Discrepancy                                    & needs;could've                                & 33.684                   & 29.047                     & 18.622                   & 17.041                     \\
Tentative                                      & nearly;disorient                              & 121.143                  & 101.765                    & 57.549                   & 48.100                     \\
Certainty                                      & factual;accuracy                                & 30.834                   & 26.742                     & 16.213                   & 15.116                     \\
Differentiation                                & nor;respective                                & 90.079                   & 79.409                     & 43.286                   & 41.288                     \\
Perceptual processes                           & bittersweet;butter                            & 48.846                   & 42.232                     & 26.355                   & 25.144                     \\
See                                            & shine;selfie                                  & 26.406                   & 21.792                     & 17.289                   & 16.141                     \\
Drives                                         & associations;orders                           & 408.641                  & 333.020                    & 212.684                  & 174.013                    \\
Affiliation                                    & socially;confiding                            & 92.258                   & 81.826                     & 48.664                   & 44.677                     \\
Achievement                                    & obtain;attain                                 & 27.246                   & 23.715                     & 15.116                   & 13.994                     \\
Power                                          & parliament;ambitions                          & 138.243                  & 110.705                    & 74.478                   & 60.190                     \\
Reward                                         & add;takes                                     & 26.496                   & 21.956                     & 15.985                   & 14.337                     \\
\rowcolor[HTML]{9AFF99} 
Past Focus                                     & appeared;sung                                 & 149.793                  & 139.685                    & 89.218                   & 80.806                     \\
Present Focus                                  & believe;cashes                                & 861.383                  & 736.258                    & 416.913                  & 371.871                    \\
Future Focus                                   & i'mma;expect                                  & 26.156                   & 20.537                     & 16.197                   & 13.496                     \\
Relativity                                     & young;slowing                                 & 301.238                  & 255.114                    & 130.221                  & 113.037                    \\
Motion                                         & catch;lunge                                   & 45.904                   & 37.460                     & 24.249                   & 20.790                     \\
Space                                          & biggest;little                                   & 167.501                  & 139.688                    & 77.001                   & 65.491                     \\
Time                                           & presently;modernly                            & 92.210                   & 80.913                     & 42.411                   & 37.788                     \\
\rowcolor[HTML]{9AFF99} 
Work                                           & hiring;legals                                 & 31.227                   & 28.409                     & 21.539                   & 19.167 \\
\bottomrule
\end{supertabular*}
\end{center}

\twocolumn

\end{document}